\documentclass[runningheads]{llncs}

\makeatletter
\@ifundefined{cellcolor}{%
	\usepackage{colortbl}%
}{}
\makeatother
\usepackage{marvosym}        
\usepackage{wrapfig}
\usepackage{makecell}
\usepackage{multirow}
\usepackage{algorithm}
\usepackage{algpseudocode}
\usepackage{amsmath}
\definecolor{mygray}{gray}{0.9}  
\usepackage{eccv}

\usepackage{eccvabbrv}

\usepackage{graphicx}
\usepackage{booktabs}

\usepackage[accsupp]{axessibility}  % Improves PDF readability for those with disabilities.

\usepackage{hyperref}

\usepackage{orcidlink}

\begin{document}
	
	\title{IREU: Identity-Related Encoder-Only Unlearning for Customized Portrait Generation} 
	
	\titlerunning{IREU for Customized Portrait Generation}

	\author{Chaoyi Shi\inst{1}\orcidlink{0009-0008-5579-8212} \and
		Shanshan Zhang\inst{1,2}\textsuperscript{(\Letter)}\orcidlink{0000-0003-4013-6300} \and
		Jian Yang\inst{1,2}\orcidlink{0000-0003-4800-832X}}
	
	\authorrunning{C.~Shi et al.}
	
	
	\institute{PCA Lab, School of Computer Science and Engineering, Nanjing University of Science and Technology \and
	PCA Lab, School of Intelligence Science and Technology, Nanjing University\\
		\email{\{chaoyishi,shanshan.zhang,csjyang\}@njust.edu.cn}}
	
	\maketitle		
	\begin{abstract}
		Customized Portrait Generation (CPG) technologies have been widely used to generate high-fidelity person images given an input image indicating the identity and a text prompt indicating the required edits. Yet these methods pose significant privacy risks by spreading fake visual information. Against such risks, each public generator should be able to suppress its generation ability for a particular person when requested. Therefore, in this work we investigate the identity unlearning problem for CPG. Since there are no previous methods in this field, we propose a simple baseline that updates the image encoder by minimizing identity similarity between generated and input images for target identities to be unlearned, while maximizing it for identities to be retained. However, we find such a global perturbation in the feature space harms the fidelity of generated images for other identities to be retained. To solve this problem, we propose a novel method IREU, which first locates identity-related features in an offline manner and then only performs feature perturbations on them. The experimental results show that our proposed method IREU achieves better identity unlearning performance for target identities to be unlearned, and also keeps high fidelity for other identities to be retained. In addition, our unlearned image encoder is generalizable across different generators with the same encoder without fine-tuning, which is friendly for deployment in practice.
	\end{abstract}		
	\section{Introduction}
	\label{sec:intro}	
	In recent years, diffusion-based text-to-image models (e.g., Stable Diffusion, SDXL) have substantially advanced Customized Portrait Generation (CPG) \cite{DBLP:conf/cvpr/LiCWQCS24,DBLP:journals/ijcv/XiaoYFDH25}. With a few images or short prompts, these models synthesize high-fidelity, editable portraits and support applications such as attribute editing, identity blending, and scenario recontextualization (e.g., personalized avatars, character animation, virtual try-on \cite{Kim_2024_CVPR,DBLP:conf/eccv/TianWZB24,DBLP:conf/cvpr/Zeng0NTWL24}). However, the same convenience poses significant privacy risks, especially the unintended reproduction of real-world identities. For example, adversaries can exploit this behavior for impersonation, deepfake abuse, and loss of credibility \cite{DBLP:conf/cvpr/HuangC0023,Wang_2025_CVPR,DBLP:conf/cvpr/Kang0JC25}.

	To protect privacy, some methods primarily focus on image-level protection, such as injecting imperceptible adversarial perturbations into user images to disrupt unauthorized usage \cite{DBLP:conf/cvpr/Song0CS25,DBLP:journals/corr/abs-2504-17894}. Although this image-centric protection can achieve privacy preservation at a low cost, this is only feasible provided that all images of that identity are protected, which is clearly impractical, as malicious users can obtain unprotected images such as already circulated images and covert photography. Moreover, according to the EU's General Data Protection Regulation (GDPR) \cite{gdpr}, service providers (model owners) are obliged to implement erasure upon user requests. Therefore, to ensure robust privacy protection and regulatory compliance, the defense paradigm must shift from perturbing images to perturbing models, with model-based generative unlearning effectively addressing this requirement.
	
	\begin{figure*}[t]
		\centering
		\includegraphics[width=1\textwidth]{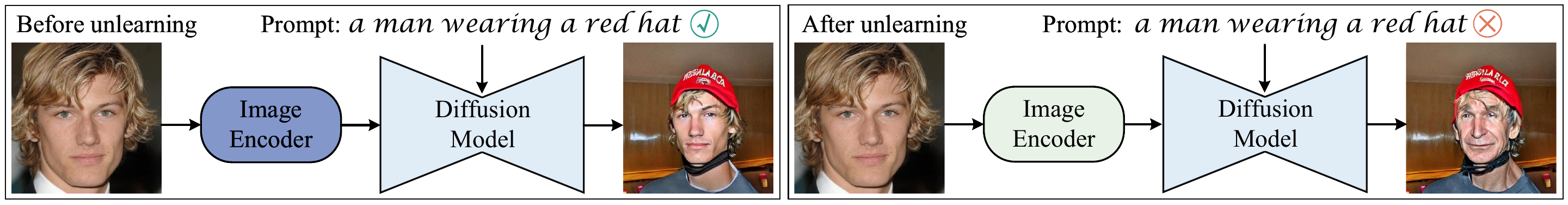} 
		\caption{
			Illustration of our identity unlearning paradigm. Left: a typical CPG pipeline, making edits based on the text prompt while keeping the same identity as the input. Right: only the image encoder is updated (w/ the diffusion model frozen) during unlearning, so that the generated images exhibit a different identity from the input.
		}
		\label{fig:1}
	\end{figure*}

	Although model-level unlearning provides a more robust solution, achieving identity unlearning for CPG remains highly challenging. Existing generative identity unlearning methods \cite{DBLP:conf/cvpr/ShaheryarLJ25,Perturb_a_Model} typically fine-tune the denoising network. However, the U-Net contains extensive generative prior knowledge, and directly erasing identity information within this high-dimensional parameter space can easily trigger catastrophic forgetting \cite{Lyu_2024_CVPR}, thereby degrading the overall fidelity of generated images. Furthermore, recent CPG pipelines (e.g., PhotoMaker \cite{DBLP:conf/cvpr/LiCWQCS24}, FastComposer \cite{DBLP:journals/ijcv/XiaoYFDH25}) consist of a pre-trained encoder and a diffusion model. Unlearning on the U-Net limits its generalization ability across generators that usually share the same encoder family but differ in denoisers. In contrast, an encoder-only design is more friendly to deployment, i.e., once the encoder is optimized towards identity unlearning, it can be plugged into an arbitrary generator that shares the same encoder. Thus, as illustrated in Fig. \ref{fig:1}, in this work we propose novel unlearning methods for CPG by modifying the encoder only.

\begin{figure*}[t]
		\centering
		\includegraphics[width=1\textwidth]{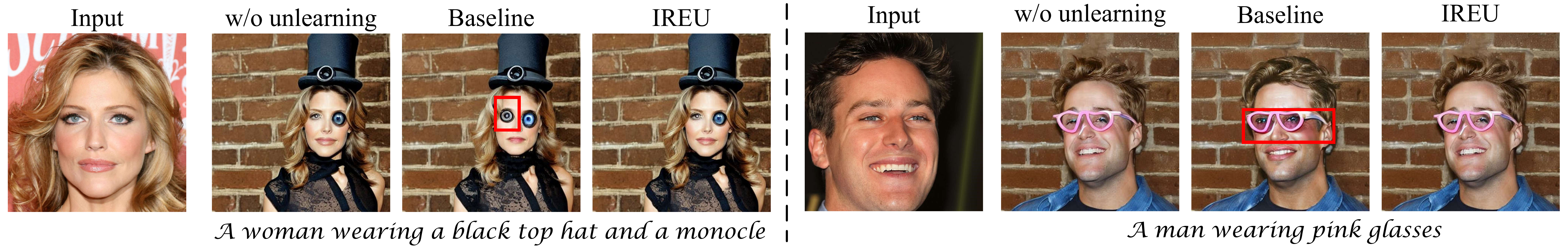} 
		\caption{
			The major limitation of the baseline method using global feature perturbation. After unlearning on target identities, the generated images for other identities to be retained are of lower fidelity, showing some noticeable discrepancies indicated by red boxes. In contrast, our method IREU addresses this problem by introducing identity-related feature perturbation.
		}
		\label{fig:2}
	\end{figure*}

	Following this idea, we first establish a simple baseline, where the encoder is optimized by minimizing the identity similarity between each target identity to be unlearned and its corresponding customized generated image, and maximizing this identity similarity for other identities. 
	However, although the target identities can be well unlearned, this kind of global perturbation in the feature space is harmful for keeping high fidelity for other identities to be retained, leading to noticeable quality degradation, as shown in Fig.~\ref{fig:2}.
	To address this problem, we propose to perform perturbation only on identity-related features, so as to suppress the target identity in the generated images while largely alleviating the negative effect on fidelity for other identities in the meantime.
	
	Building on this insight, we propose a novel \textbf{I}dentity-\textbf{R}elated \textbf{E}ncoder-Only \textbf{U}nlearning (\textbf{IREU}) framework, which performs all updates in the image feature space while keeping the diffusion model frozen.
	To isolate identity-related features, we first build Face-Swap pairs that differ in identity and compute their embedding differences. Dimensions with the largest deltas with respect to identity are selected as identity-related features. We then apply localized perturbations along these dimensions on the inputs to be unlearned, shifting their representations away from the original identity. For other identities to be retained, we follow the baseline’s practice and keep consistency between the unlearned and original encoder to keep features unchanged.
	
	Our contributions are summarized as follows:
	\begin{itemize}
		\item To the best of our knowledge, we are the first to address the identity unlearning problem specifically for CPG by updating the image encoder only, making it generalize well across different generators.
		\item In order to obtain high identity dissimilarity for target IDs to be unlearned and to preserve high fidelity for other IDs to be retained, we propose to locate identity-related features in the embedding space via Face-Swap and then apply localized perturbations via identity-related transformation.
		\item Extensive experiments show that our method outperforms recent identity unlearning approach and achieves a better trade-off between unlearning and retaining. The unlearned encoder is plug-and-play across diverse generators and preserves fidelity without additional training.
	\end{itemize}
	
	\section{Related Work}
	
	\subsection{Customized Portrait Generation and Privacy Issues}
	Text-to-image diffusion models have rapidly advanced the landscape of Customized Portrait Generation (CPG). Early representative methods \cite{DBLP:conf/cvpr/RuizLJPRA23,DBLP:conf/iclr/GalAAPBCC23,voynov2023P+} can reproduce a subject’s appearance from only a few input images, yet typically incur long finetuning time. Subsequent parameter-efficient adaptations \cite{Li_HyperLoRA_2025_CVPR,DBLP:conf/cvpr/LiCWQCS24,DBLP:journals/ijcv/XiaoYFDH25,wang2024instantid} further accelerate personalization and improve identity consistency.
	
	At the same time, the high fidelity and accessibility of CPG heighten privacy risks because outputs are tightly coupled with real humans. Existing protections mainly act on the input–output boundary: provenance methods inject robust watermarks for attribution \cite{feng2024aqualora,SleeperMark_2025_CVPR}, input-side defenses add imperceptible perturbations to hinder unauthorized editing \cite{yang2025variance,PID}, and privacy-friendly generation schemes reduce identifiability via domain constraints \cite{Shamshad_2023_CVPR}. While useful, these approaches do not selectively remove the model’s internal capacity to regenerate a specific identity. We instead pursue a model-internal method: IREU converts a Face-Swap identity direction into a masked, identity-related perturbation while keeping the diffusion model frozen, thereby suppressing the target identity, preserving fidelity, and transferring across CPG pipelines without extra training.

	\subsection{Unlearning in Diffusion Models}
	Machine unlearning was first established in image classification \cite{DBLP:journals/tnn/LiZGCZKF25,DBLP:conf/cvpr/ChenGL0W23,DBLP:conf/cvpr/SpartalisSGD25,DBLP:conf/cvpr/KhalilBL0B025,DBLP:conf/cvpr/Zhong000ZLL25,Zhou_2025_CVPR,DBLP:conf/cvpr/AhmedBRDKNGR25}, aiming to remove the influence of targeted data with minimal model changes \cite{DBLP:journals/tnn/LiZGCZKF25}. As demands for controllability grew, attention shifted to representation-level editing in generative models, exploring concept erasure and generative unlearning as internal mechanisms for selective removal \cite{Lee_2025_CVPR,DBLP:conf/cvpr/ChoiCLSN22}.  In diffusion models, early concept erasure typically involves small-scale finetuning of the denoising network to suppress specified concepts \cite{Gandikota_2023_ICCV}, whereas closed-form parameter editing adjusts the model's parameters through finetuning to enable efficient cross-attention edits without full retraining \cite{Gandikota_2024_WACV,Lu_2024_CVPR}.  Machine unlearning has been applied not only to remove specific styles or unsafe content \cite{DBLP:conf/aaai/WuZYWCZHZ025,DBLP:conf/cvpr/00210HH25,NEURIPS2024_40954ac1,Zhang_2024_CVPR,10.1007/978-3-031-73668-1_5,Gao_mmu_2025_ICCV}, but also to selectively suppress fine-grained attributes such as hairstyles and hats \cite{DBLP:conf/aaai/MoonC024,NEURIPS2024_2b09bb02}.
	
	Closer to our setting is identity-level generative unlearning. GUIDE \cite{DBLP:conf/cvpr/SeoLLMP24} removes an identity’s generative capability from a pre-trained GAN given a single image, while BIA \cite{DBLP:conf/cvpr/ShaheryarLJ25} builds a black-hole-like absorption in the diffusion UNet bottleneck to absorb identity representations. However, GUIDE is tied to GANs whereas recent CPG pipelines are predominantly diffusion-based, and BIA’s UNet-bottleneck editing tightly couples unlearning to a specific generator, making transfer across CPG costly. APDM \cite{Perturb_a_Model} adjusts model parameters specifically for fine-tuning-based CPG methods (e.g., DreamBooth), which results in significant time consumption and renders it inapplicable to current mainstream CPG approaches. In contrast, we perform unlearning purely in the encoder embedding space while freezing the diffusion backbone, enabling training-free transfer across different CPG pipelines that share the same encoder.
	\begin{figure*}[t]
		\centering
		\includegraphics[width=1\textwidth]{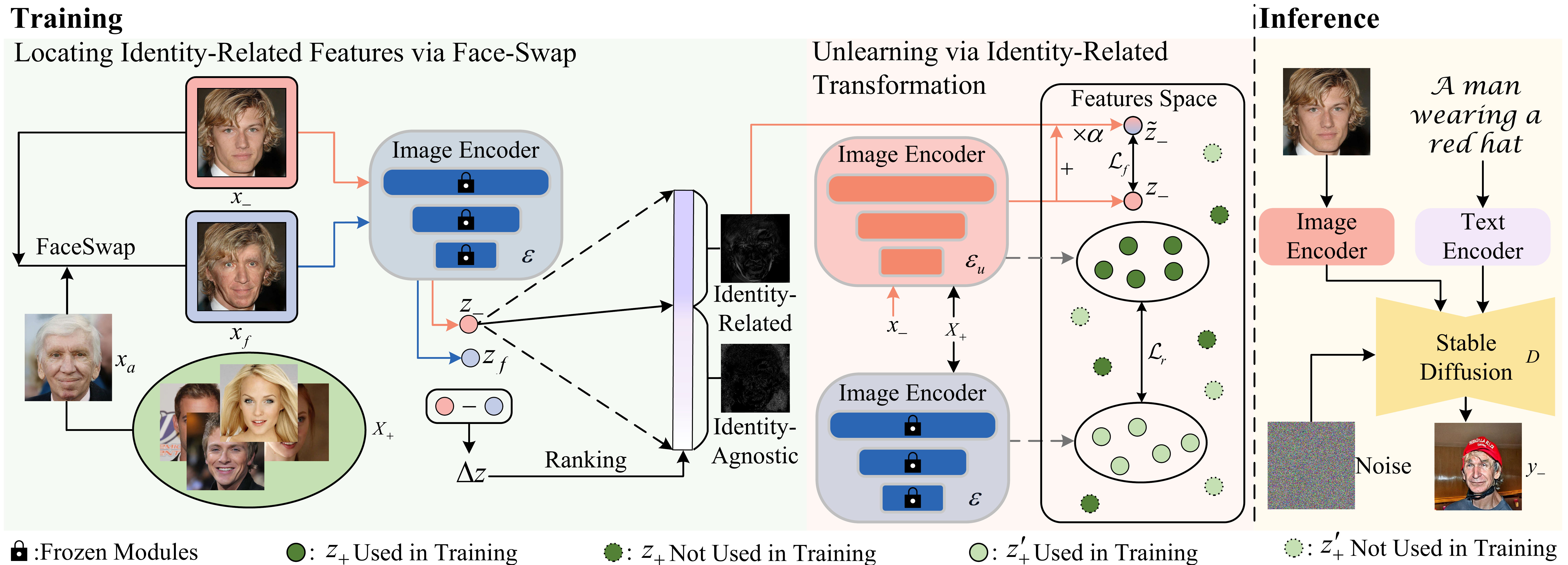} 
		\caption{
			Overview of our proposed method IREU, which consists of two key components: (1) \textbf{Locating Identity-Related Features:} given an image $x_{-}$ to be unlearned and its Face-Swap counterpart $x_{a}$, we encode them with the original image encoder $\varepsilon$ to obtain their embeddings. Then we compute the difference $\Delta z$, and rank $|\Delta z|$ to build a binary mask $\mathcal{M}$ that selects identity-related dimensions for feature perturbation. (2) \textbf{Unlearning via Identity-Related Transformation:} we construct a virtual identity $\tilde{z}_{-}$ with a magnitude $\alpha$, and update the unlearning encoder $\varepsilon_{u}$ through an unlearning loss $\mathcal{L}_{f}$ to minimize the distance between ${z}_{-}$ and $\tilde{z}_{-}$. For other identities to be retained, we employ a retaining loss $\mathcal{L}_{r}$ to minimize the output discrepancy between $\varepsilon_{u}$ and $\varepsilon$. Notably, $\mathcal{L}_{r}$ is computed using a randomly sampled image to be retained during each iteration. 
			During inference, the unlearned encoder suppresses the target identity to be unlearned while maintaining the fidelity of other identities to be retained.
		}
		\label{fig:overview}
	\end{figure*}
	
	\section{Method}
	
	\subsection{Preliminaries}
	\label{sce:prel}
	\paragraph{CPG pipeline.}
	The state-of-the-art customized portrait generators follow a standard text-to-image diffusion model that synthesizes personalized portraits by conditioning on a customized text prompt and one or more input images. 
	Both image and text encoders are inherited from CLIP~\cite{DBLP:conf/icml/RadfordKHRGASAM21}. 
	The text prompt provides high-level semantics and intent, and its CLIP text embedding modulates the denoising network via cross-attention. The input images supply person-specific appearance cues, and their CLIP image embeddings anchor the generated face to the original identity while keeping pose and background editable. Given an input image $x$, we obtain its feature vector $z$ through the image encoder $\varepsilon$, i.e., $z = \varepsilon(x)$.
	A conditional Stable Diffusion model $D$ then generates $y = D(z, t)$ based on the image embedding $z$ and text prompt $t$.
	
	\textbf{Identity Unlearning in CPG.}
	Let $\mathcal{I}$ be the set of all identities. We split it into the unlearning subset $\mathcal{I}_{-}$ (target identities to be unlearned) and the retaining subset $\mathcal{I}_{+}=\mathcal{I}\setminus\mathcal{I}_{-}$ (other identities to be retained).
	For each identity $i\in\mathcal{I}$, let $X(i)=\{x_i^{j}\}_{j=1}^{n_i}$ be its image set with $n_i\ge 1$. The entire image dataset is noted as \(X = X_- \cup X_{+}\) with \(X_- \cap X_{+}=\varnothing\), where $
	X_{-}=\bigcup_{i\in\mathcal{I}_{-}} X(i)$ and $
	X_{+}=\bigcup_{i\in\mathcal{I}_{+}} X(i)$. 
	In the context of CPG, the goal is to ensure that for an image $x_{-}^j \in X_-$, the image embedding $z_-^j= \varepsilon_u(x_-^j)$ produced by the post-unlearning image encoder $\varepsilon_u$ leads to a generated image $y_-^j = D(z_-^j, t)$ that no longer retains the identity characteristics consistent with $x_{-}^j$.
	
	For other identities to be retained, we randomly sample an image $x_{+}^j\in X_{+}$, and the post-unlearning image encoder $\varepsilon_u$ should make sure that the generated image $y_+^j = D(\varepsilon_u(x_+^j), t)$ retains the same identity as $x_{+}^j$.
	
	For target identities to be unlearned, prior work typically provides the model with one input image per identity~\cite{DBLP:conf/cvpr/SeoLLMP24,DBLP:conf/cvpr/ShaheryarLJ25}. We follow this one-shot setting with $n_i=1$ and, for clarity, omit superscripts in $x_-^{j}$ and $x_+^{j}$, writing them simply as $x_-$ and $x_+$ in what follows.
	
	\subsection{Overview}
	
	Since there are no previous methods designed for identity unlearning in CPG, we first establish a \textit{simple baseline} method. 
	The image encoder is updated with two objectives: for an image from any target identity to be unlearned ($x_{-}$), the $\ell_2$ distance between the outputs of the post-unlearning encoder $\varepsilon_u$ and the original encoder $\varepsilon$ is maximized, pushing its representations away from the original features; for an arbitrary image to be retained $x_{+} \in X_{+}$, the $\ell_2$ distance between the outputs of the post-unlearning encoder $\varepsilon_u$ and the original encoder $\varepsilon$ is minimized, keeping the representations unchanged. In this way, the entire feature space for each image is jointly perturbed during the optimization process.
	
	Since global feature perturbation hurts the fidelity of generated images, we propose to perform the perturbation only on identity-related features instead of the entire feature vector. To achieve this goal, we first need to locate those identity-related features.
	Specifically, we construct Face-Swap pairs in which only the identity differs between the two images, and we treat the elements with the largest differences as identity-related features.
	
	After that, we perturb identity-related features for the input images to be unlearned so as to change the identity in the feature space; while for other identities to be retained, we minimize the features from the unlearned and original encoders, just as the baseline does. 
	
	\subsection{Locating Identity-Related Features via Face-Swap}
	To localize identity-related features in the embedding space, we use Face-Swap \cite{DBLP:conf/cvpr/GaoHFLH21} to synthesize a surrogate face for the target identity to be unlearned and compute the latent difference between the original and the swapped image. We then decouple the entire feature vector into identity-related and identity-agnostic components, and only perform perturbation on the former group.
	\begin{figure*}[t]
		\centering
		\includegraphics[width=1\textwidth]{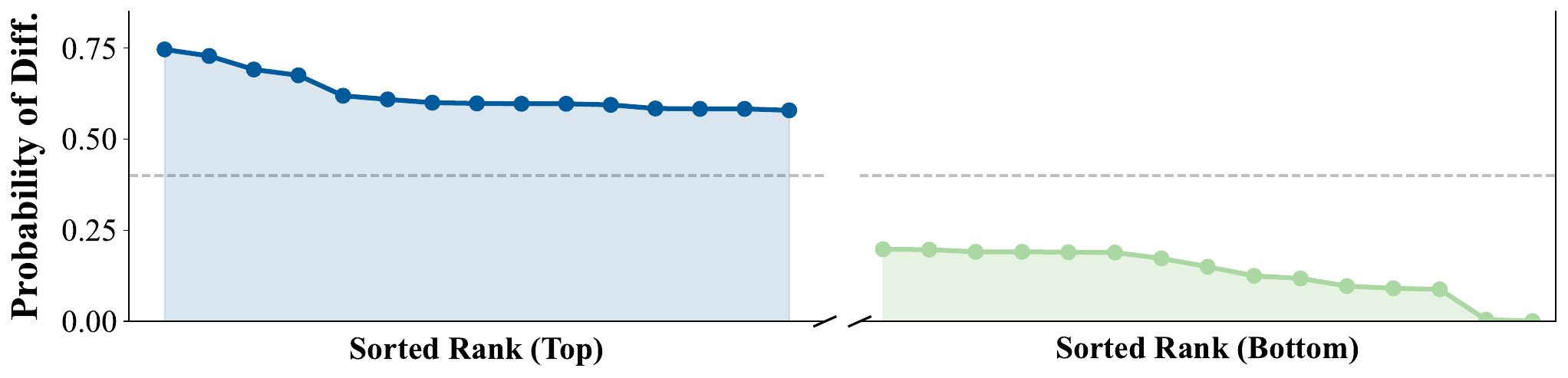} 
		\caption{Distribution of difference features across 1,000 Face-Swap pairs. The non-uniform distribution exhibits a clear bias towards specific dimensions, which motivates our dimensional masking strategy to isolate identity-related features.}
		\label{fig:mov}
	\end{figure*}
	Specifically, as shown in Fig. \ref{fig:overview}, for an image from a target identity to be unlearned $x_{-}$, we obtain its swapped counterpart $x_{f} = F(x_{-}, x_{a})$, where $x_{a}$ is randomly sampled from the other identity image set $X_+$, and $F(\cdot)$ is the Face-Swap function. This function generates an image that maintains visually consistent context with $x_{-}$ but represents the identity of $x_{a}$. We treat $F(\cdot)$ as an off-the-shelf, non-differentiable operator used only to construct Face-Swap pairs, where gradients do not flow through $F$.	
	
	After the identity shift between \(x_-\) and \(x_f\), we compute the embedding difference as follows:
	\begin{equation}
		\label{deltaz}
		{ \Delta z={{z}_{-}}-{{z}_{f}}},
	\end{equation}
	where ${z}_{-}=\varepsilon(x_-)$, ${z}_{f}=\varepsilon(x_f)$ and the vector $\Delta z$ captures the change in embedding induced by identity shift.
	
	In the difference vector $\Delta z$, the absolute value of each component reflects how strongly that dimension responds to the identity change: a larger $|\Delta z_i|$ implies a higher identity sensitivity. Additionally, we conduct statistics on 1,000 Face-Swap image pairs and calculate the occurrence probability of the top 40\% of $|\Delta z_i|$ values. As shown in Fig. \ref{fig:mov}, the differences are not uniformly distributed across all dimensions but exhibit a clear bias. This observation motivates us to employ a dimensional masking strategy from a new perspective to locate identity-related features. We therefore sort all dimensions in descending order by $|\Delta z_i|$ and select the top dimension entries as identity-related, while regarding the remaining low-magnitude dimensions as approximately identity-agnostic.	
	
	We set a quantitative threshold $k$ to construct a mask $\mathcal{M}\in {{\left\{ 0,1 \right\}}^{d}}$, which will be used to isolate identity-related features while suppressing identity-agnostic ones:
	\begin{equation}
		\label{m}
		\mathcal{M}_{}=\mathbb{I}(\operatorname{rank}\big(|\Delta z|_{[i]}\big)\le \lfloor k\cdot d \rfloor),
	\end{equation}
	where $\operatorname{rank}(|\Delta z|_{[i]})$ denotes the descending order index of the $i$-th entry when sorting the absolute values $|\Delta z|$ across all $d$ dimensions, so that both positive and negative identity shifts are treated equally; $k\in(0,1]$ is the kept ratio; $d$ is the length of the embedding vector; $\mathbb{I}(\cdot)$ is the indicator function that returns 1 if the condition is true and 0 otherwise. 
	
	\subsection{Unlearning via Identity-Related Transformation}
	Leveraging the identity-related features selected by the mask $\mathcal{M}$, we only modify the identity-related elements while leaving identity-agnostic features unchanged. This ensures the protection of retained identities and the preservation of the original distribution.
	This module produces the virtual identity $\tilde{z}_{-}$ as follows:
	\begin{equation}
		\label{zu}
		{ {{\tilde{z}}_{-}}={{z}_{-}}+\alpha (\mathcal{M}\odot \Delta z)},
	\end{equation}
	where $\odot$ denotes element-wise multiplication, and $\alpha$ is the magnitude coefficient. Compared to the baseline, which maximizes the distance to the original encoder along all dimensions, this construction restricts the shift of $z_{-}$ to identity-related dimensions and avoids unnecessary changes to identity-agnostic features.
	\begin{table*}[t]
		\begin{center}
			\caption{Comparison of different methods on FastComposer~\cite{DBLP:journals/ijcv/XiaoYFDH25} w.r.t. identity-centric metrics under both single-identity (oneID) and multi-identity (fiveID) settings. 
				We re-implement BIA~\cite{DBLP:conf/cvpr/ShaheryarLJ25} based on Stable Diffusion and plug it in the same generator for a fair comparison.
				Best results are in \textbf{bold}.}	
			
			\setlength{\tabcolsep}{0.35mm}
			\footnotesize

			\begin{tabular}{c|c|cccc|cc|c} 
				\toprule[1pt]
				\multirow{2}{*}{ID Type} & \multirow{2}{*}{Methods} & \multicolumn{4}{c|}{Target ID ($\downarrow$)} & \multicolumn{2}{c|}{Other ID ($\uparrow$)} & \multirow{2}{*}{$\Delta$ID($\uparrow$)} \\
				\cline{3-8} 
				& & $\text{ID}_\text{target}^{o}$ & $\text{ID}_\text{target}^{o,\text{unseen}}$ & $\text{ID}_\text{target}^\text{human}$ & $\text{ID}_\text{target}^{g}$ & $\text{ID}_\text{other}^{o}$ & $\text{ID}_\text{other}^{g}$ & \\
				\midrule
				\multirow{3}{*}{oneID}
				&BIA~\cite{DBLP:conf/cvpr/ShaheryarLJ25}~\scriptsize{[CVPR 25']} & 0.05 & 0.18 & 27.92\% & 0.30 & 0.55 & 0.82 & 0.52 \\
				& \cellcolor{mygray}Baseline (Ours) &\cellcolor{mygray}0.03 &\cellcolor{mygray}0.16 &\cellcolor{mygray}8.33\% &\cellcolor{mygray}\textbf{0.03} &\cellcolor{mygray}0.14 &\cellcolor{mygray}0.51 &\cellcolor{mygray}0.48 \\
				
				& \cellcolor{mygray}IREU (Ours) &\cellcolor{mygray}\textbf{0.01} &\cellcolor{mygray}\textbf{0.15} &\cellcolor{mygray}\textbf{6.95\%} &\cellcolor{mygray}0.21 &\cellcolor{mygray}\textbf{0.55} &\cellcolor{mygray}\textbf{0.85} &\cellcolor{mygray}\textbf{0.64} \\
				\midrule
				\multirow{3}{*}{fiveID}
				&BIA~\cite{DBLP:conf/cvpr/ShaheryarLJ25}~\scriptsize{[CVPR 25']} & 0.15 & 0.32 & 59.03\% & 0.43 & 0.52 & 0.65 & 0.22 \\
				& \cellcolor{mygray}Baseline (Ours) & \cellcolor{mygray}0.02 & \cellcolor{mygray}0.19 & \cellcolor{mygray}\textbf{11.81\%} & \cellcolor{mygray}\textbf{0.13} & \cellcolor{mygray}0.01 & \cellcolor{mygray}0.20 & \cellcolor{mygray}0.07 \\
				
				& \cellcolor{mygray}IREU (Ours) & \cellcolor{mygray}\textbf{0.02} & \cellcolor{mygray}\textbf{0.15} & \cellcolor{mygray}12.50\% & \cellcolor{mygray}0.23 & \cellcolor{mygray}\textbf{0.53} & \cellcolor{mygray}\textbf{0.78} & \cellcolor{mygray}\textbf{0.55} \\
				\bottomrule[1pt]
			\end{tabular}	
			
			\label{table:1}
		\end{center}
	\end{table*}
	We then take $\tilde{z}_{-}$ as an explicit reference for updating $\varepsilon_u$ on the target identity to be unlearned. For an image from the target to be unlearned $x_- \in X_{-}$, we encode it with the trainable encoder to obtain $z_- = \varepsilon_u(x_-)$. We optimize the image encoder by minimizing the following $\ell_2$ distance:
	\begin{equation}
		\label{eq:loss_u}
		\mathcal{L}_{f}
		= \bigl\| z_- - {{\tilde{z}}_-} \bigr\|_2 .
	\end{equation}
	
	This unlearning loss allows the final features to deviate largely from the original features in terms of identity, so that the generated images look dissimilar from the original identity while preserving high fidelity.
	
	For other identities to be retained, we reuse the retaining loss from the baseline:
	\begin{equation}
		\label{eq:loss_r}
		\mathcal{L}_{r}
		= \bigl\| z_{+} - z'_{+} \bigr\|_2 ,
	\end{equation}
	where $z_{+} = \varepsilon_u(x_{+})$ and $z'_{+} = \varepsilon(x_{+})$. This loss keeps the features aligned with the original ones for other identities to be retained.
	
	Combining Eqs. \ref{eq:loss_u} and \ref{eq:loss_r}, the final training objective simply balances the unlearning and retaining terms:
	\begin{equation}
		\label{eq:final_obj}
		\mathcal{L}_u
		= \lambda_{1} \mathcal{L}_{f}
		+  \mathcal{L}_{r},
	\end{equation}
	where $\lambda_1$ is a hyperparameter that controls the weight of the loss function.
	
	We optimize only the image encoder $\varepsilon_u$ across the entire CPG pipeline, keeping all other components frozen.
	
	\section{Experiments}
	\begin{figure*}[t]
		\centering
		\includegraphics[width=0.95\textwidth]{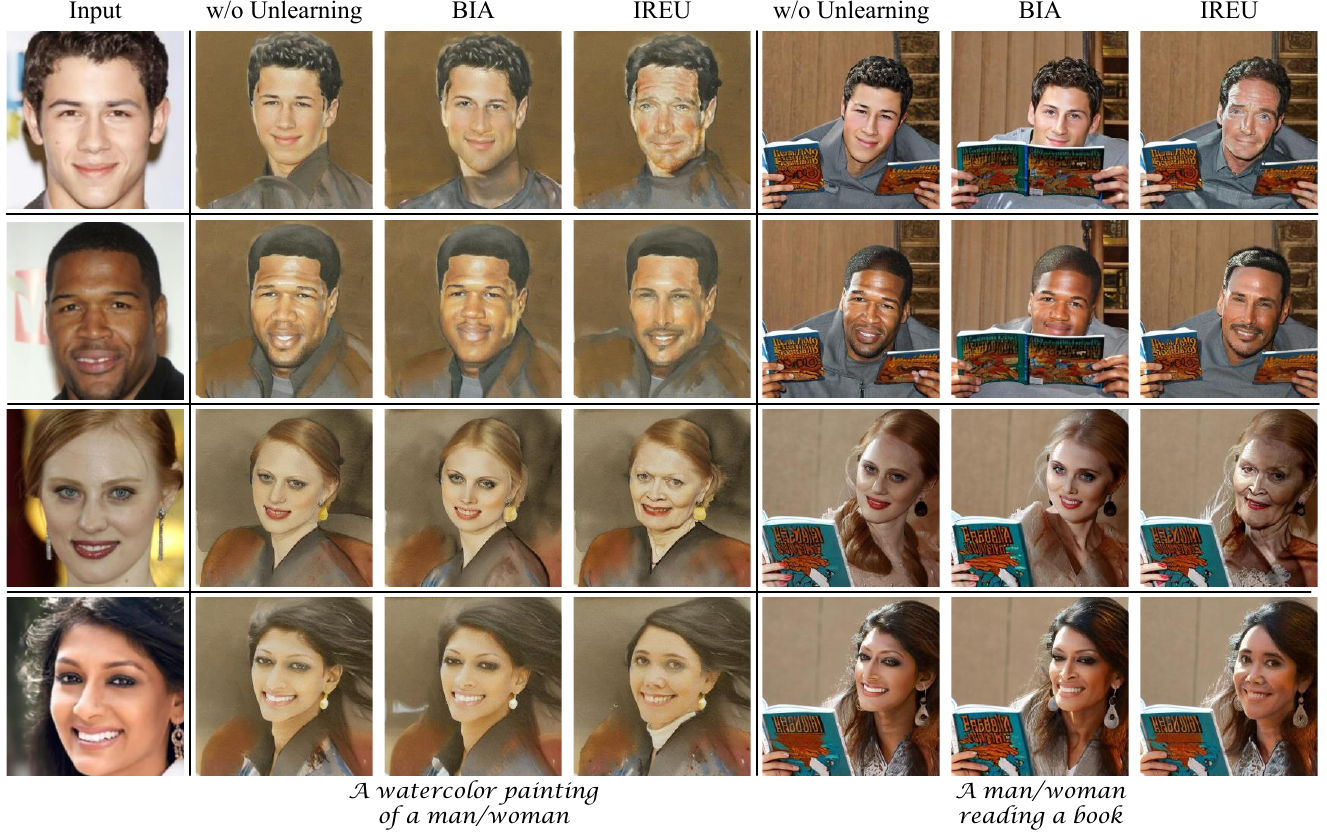}
		\caption{Comparison of visual quality with BIA~\cite{DBLP:conf/cvpr/ShaheryarLJ25} for unlearning under the oneID setting on FastComposer~\cite{DBLP:journals/ijcv/XiaoYFDH25}. For each input image (first column), both BIA and our method perform identity unlearning from the same original model. Text prompts are included below each generated image.}
		\label{fig:ulone}
	\end{figure*}

	\subsection{Implementation Details}
	We use CelebA-HQ~\cite{DBLP:conf/iclr/KarrasALL18} as the primary face dataset and split it into training, validation, and test sets with a 7:1:2 ratio. Rather than consuming the full training split, we randomly sample training images according to the number of iterations. To evaluate unlearning performance on unseen images of the same identity, we additionally use CelebA~\cite{DBLP:conf/iccv/LiuLWT15}. Our CPG pipeline utilizes FastComposer~\cite{DBLP:journals/ijcv/XiaoYFDH25} for single-subject generation with a resolution of 512$\times$512, employing Stable Diffusion v1.5 as its backbone and CLIP ViT-L/14 as the image encoder. To evaluate transferability, we directly plug the unlearned encoder into PhotoMaker~\cite{DBLP:conf/cvpr/LiCWQCS24} and the multi-subject FastComposer setting without any additional training.
	
	The framework is implemented in PyTorch and optimized with Adam at a learning rate of \(1\times10^{-5}\). All experiments are conducted on a single NVIDIA GeForce RTX 3090 GPU with a batch size of 1. For the single-identity (oneID) unlearning task, we train for 1{,}000 iterations. In addition, we construct an extended multi-identity (fiveID) unlearning task, for which we train for 12{,}000 iterations. For inference, we employ 50 denoising steps to generate customized images. The hyperparameters are set as follows: \(k=0.4\), \(\lambda_{1}=0.1\), and $\alpha =100$.
	
	\subsection{Evaluation Metrics}
	\textbf{Identity similarity.}
	To evaluate the effectiveness of our method, we report a set of identity-centric metrics. We employ CurricularFace \cite{DBLP:conf/cvpr/HuangWT0SLLH20} to measure the identity similarity between two images: $s(a,b)$. Let $X_-^{\text{train}}$ be the set of training images of identities to be unlearned, $X_-^{\text{unseen}}$ a disjoint set of unseen images from the same identities.
	We report:\begin{itemize}
		\item $\text{ID}_\text{target}^{o}$~($\downarrow$): $\mathbb{E}_{x_-\in {X_-^{\text{train}}}}\, s\!\left(y_-, x_-\right)$,  the average identity similarity between the generated image and the input image to be unlearned.
		\item $\text{ID}_\text{target}^{o,\text{unseen}}$~($\downarrow$): $\mathbb{E}_{x_-\in {X_-^{\text{unseen}}}}\, s\!\left(y_-,\,x_-\right)$, the average similarity between the generated images and \textit{unseen} images of the same target identity.
		\item $\text{ID}_\text{target}^{g}$~($\downarrow$): $\mathbb{E}_{x_-\in {X_-^{\text{train}}}}\,s\!\left(y_-,\,D(\varepsilon(x_-), t)\right)$, the average identity similarity between the post-unlearning and pre-unlearning generated images under the same prompt.
		\item $\text{ID}_\text{target}^\text{human} $ ~($\downarrow$): Following the evaluation protocol of BIA, we recruited 54 participants to assess whether 12 image pairs (before and after unlearning) belong to the same identity to be unlearned.	
		\item $\text{ID}_\text{other}^{o}$~($\uparrow$): $\mathbb{E}_{x_+\in {X_+}}\,s\!\left(y_+,x_+\right)$,
		the average identity similarity between the generated image and the other identity to be retained.
		\item $\text{ID}_\text{other}^{g}$~($\uparrow$): $\mathbb{E}_{x_+\in {X_+}}\,s\!\left(y_+,\,D(\varepsilon(x_+), t)\right)$,
		the average identity similarity between the post-unlearning and pre-unlearning generated images under the same prompt.
		\item $\Delta\text{ID}$~($\uparrow$): $\text{ID}_{\text{other}}^{g} - \text{ID}_{\text{target}}^{g}$,
		the gap between retaining and unlearning in the metric space.
	\end{itemize}
	Note that $\text{ID}_{\text{target}}^{g}$, $\text{ID}_{\text{other}}^{g}$, and $\Delta\text{ID}$ are proposed in this work as CPG-specific metrics to jointly assess unlearning and retaining in the generation space.
	
	\textbf{Fidelity.}
	We report $\Delta\text{FID}$, i.e., the change in Fréchet Inception Distance (FID) score \cite{NIPS2017_8a1d6947} on the set of other identities to be retained, comparing before and after unlearning against real CelebA-HQ images. We also report PSNR, SSIM, and LPIPS to quantify perceptual and structural fidelity.
	\begin{figure}[t]
		\begin{minipage}{0.47\linewidth}
			\centering
			\includegraphics[width=\linewidth]{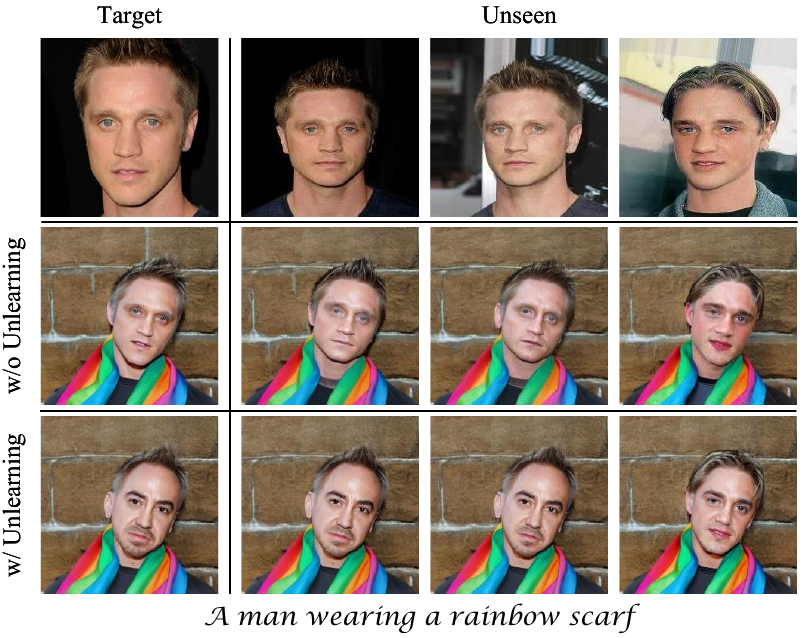}
			\captionof{figure}{Visualization results for unseen images of the same target identity to be unlearned under the fiveID setting on FastComposer~\cite{DBLP:journals/ijcv/XiaoYFDH25}. ``Target'' refers to the images used during unlearning, while ``Unseen'' denotes the images belonging to the same target identity to be unlearned but not used during unlearning.}
			\label{fig:other}
		\end{minipage}
		\hfill
		\begin{minipage}{0.45\linewidth}
			\centering
			\includegraphics[width=\linewidth]{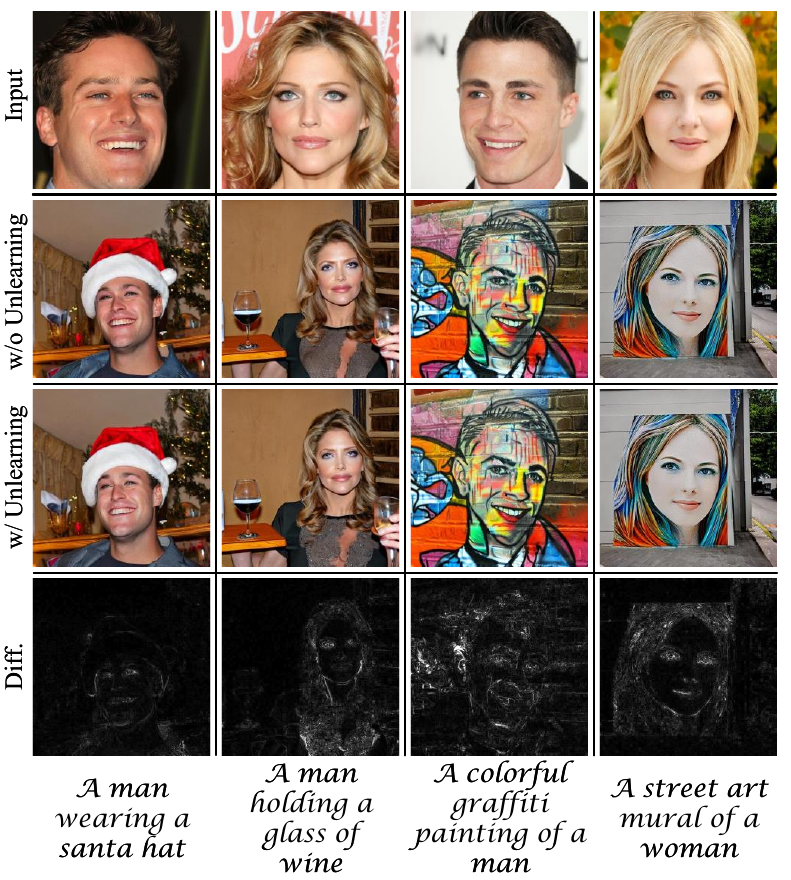}
			\captionof{figure}{Image generation quality for other identities to be retained under the fiveID setting. ``Diff.'' indicates the normalized difference between images before and after unlearning.}
			\label{fig:glob}
		\end{minipage}
	\end{figure}
	
	\begin{table}
		\caption{
			Comparison of different methods on FastComposer~\cite{DBLP:journals/ijcv/XiaoYFDH25} w.r.t. fidelity-centric metrics for other identities to be retained under single-identity (oneID) and multi-identity (fiveID) settings. }
		\label{tab:2}
		\centering
		\setlength{\tabcolsep}{3pt}
		\begin{tabular}{c|c|cccc}
			\toprule
			ID Type & Methods & PSNR($\uparrow$) & SSIM($\uparrow$) & LPIPS($\downarrow$) & $\Delta\text{FID}$($\downarrow$)\\
			\midrule
			\multirow{3}{*}{oneID}
			& BIA~\cite{DBLP:conf/cvpr/ShaheryarLJ25}~\scriptsize{[CVPR 25']} & 20.51 & 0.70 & 0.19 & 4.14\\
			& \cellcolor{mygray}Baseline & \cellcolor{mygray}22.82 & \cellcolor{mygray}0.79 & \cellcolor{mygray}0.11 & \cellcolor{mygray}4.33\\
			& \cellcolor{mygray}IREU & \cellcolor{mygray}\textbf{28.88} & \cellcolor{mygray}\textbf{0.92} & \cellcolor{mygray}\textbf{0.04} & \cellcolor{mygray}\textbf{1.02}\\
			\midrule
			\multirow{3}{*}{fiveID}
			& BIA~\cite{DBLP:conf/cvpr/ShaheryarLJ25}~\scriptsize{[CVPR 25']} & 19.46 & 0.66 & 0.22 & 2.76\\
			& \cellcolor{mygray}Baseline & \cellcolor{mygray}20.16 & \cellcolor{mygray}0.70 & \cellcolor{mygray}0.19 & \cellcolor{mygray}28.37\\
			& \cellcolor{mygray}IREU & \cellcolor{mygray}\textbf{28.25} & \cellcolor{mygray}\textbf{0.91} & \cellcolor{mygray}\textbf{0.04} & \cellcolor{mygray}\textbf{1.59}\\
			\bottomrule
		\end{tabular}
	\end{table}
	\subsection{Unlearning Performance on Target Identities}
	Table~\ref{table:1} reports unlearning performance w.r.t. identity-centric metrics under both oneID and fiveID settings. Based on the results, we have the following observations:
	(1) Our methods (including baseline and IREU) obtain better unlearning performance (lower ID similarities) than BIA across both settings and four metrics, indicating our proposed encoder-only unlearning strategy is effective.
	(2) Our method IREU obtains comparable performance for target identities to be unlearned, but achieves a better trade-off between unlearning and retaining (higher $\Delta\text{ID}$), demonstrating the impact of our proposed unlearning method focusing on identity-related features.
	(3) Our methods better handle multi-identity settings. For example, the gap w.r.t. $\text{ID}_\text{target}^\text{human}$ grows from around 20 pp (oneID) to 48 pp (fiveID).
	(4) Our methods are more robust than BIA when the generator meets unseen images from the same target identity. Our numbers w.r.t. $\text{ID}_\text{target}^{o,\text{unseen}}$ remain at 0.15, while those of BIA increase from 0.18 to 0.32.
	
	Fig.~\ref{fig:ulone} qualitatively compares IREU and BIA using the same input images and prompts. For BIA, the generated images after unlearning still look rather similar to the ones w/o unlearning, while our method IREU suppresses the identity information more thoroughly. These results indicate that our method IREU better prevents the target identity images from being edited.
	Furthermore, we show some results of unseen images (not used during unlearning) belonging to the same target identity in Fig.~\ref{fig:other}. IREU achieves identity unlearning even for unseen images. These results reveal that IREU also successfully pushes them away from the target identities in the feature space. This advantage well fulfills the requirements in real-world applications where generators usually meet unseen images.

	\subsection{Retaining Performance on Other Identities}
	For other identities to be retained, we evaluate both identity consistency ($\text{ID}_{\text{other}}^{o}$, $\text{ID}_{\text{other}}^{g}$ in Table~\ref{table:1}) and perceptual fidelity (Table~\ref{tab:2}).
	Based on the results, we have the following observations:
	(1) Our method IREU obtains the highest identity similarity and fidelity across both oneID and fiveID settings, outperforming both BIA and the baseline. These results demonstrate that our method not only preserves the identity information, but also maintains high-quality image generation for those identities to be retained, not affected by the unlearning procedure. 
	(2) The baseline fails dramatically under the fiveID setting (0.01 w.r.t. $\text{ID}_\text{other}^{o}$), indicating the global perturbation in the feature space becomes more harmful when the number of target identities to be unlearned grows. In contrast, our proposed identity-related feature perturbation overcomes this problem and thus achieves better retaining performance.
	
	In Fig.~\ref{fig:glob}, we visualize some generation results for the other identities to be retained. It can be seen that the post-unlearning outputs are closely similar to the original ones before unlearning, with the differences being barely perceptible to the human eye and mainly concentrated in high-frequency regions (e.g., hair, fine textures), as shown in the last row. This comparison indicates that our method IREU keeps the feature distribution stable and confines residual updates to visually tolerant regions, thereby avoiding global drift.
	\begin{figure}[t]
		\centering	\includegraphics[width=0.9\columnwidth]{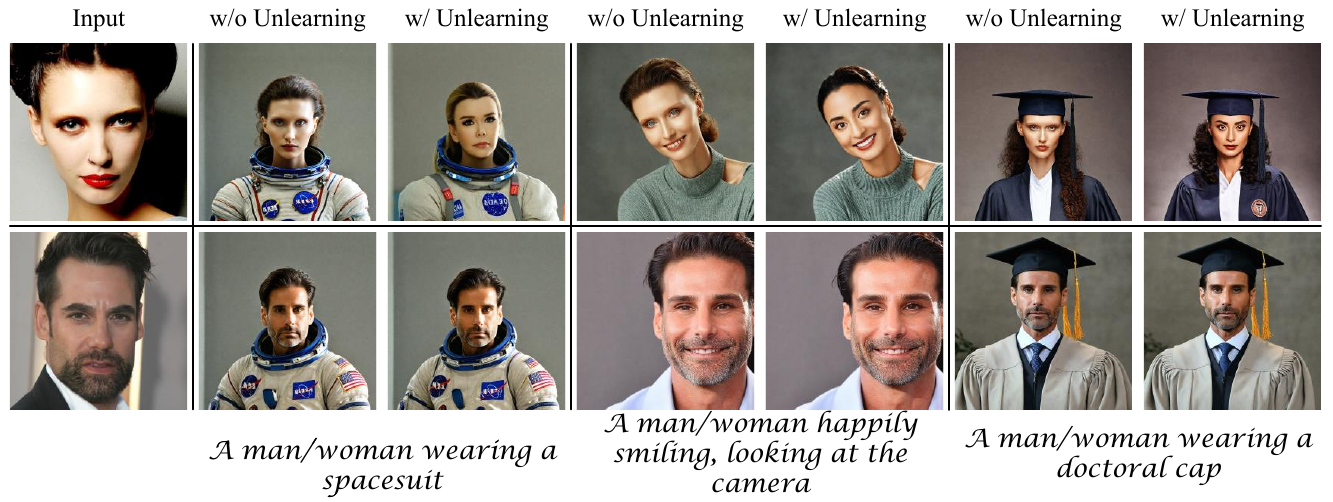} 
		\caption{  Qualitative results when our unlearned image encoder is plugged into a new generator PhotoMaker~\cite{DBLP:conf/cvpr/LiCWQCS24} without any finetuning under the fiveID setting. Top: an identity to be retained; bottom: a target identity to be unlearned.}
		\label{fig:photo}
	\end{figure}
	\begin{figure}[t]
		\centering
		\includegraphics[width=0.9\columnwidth]{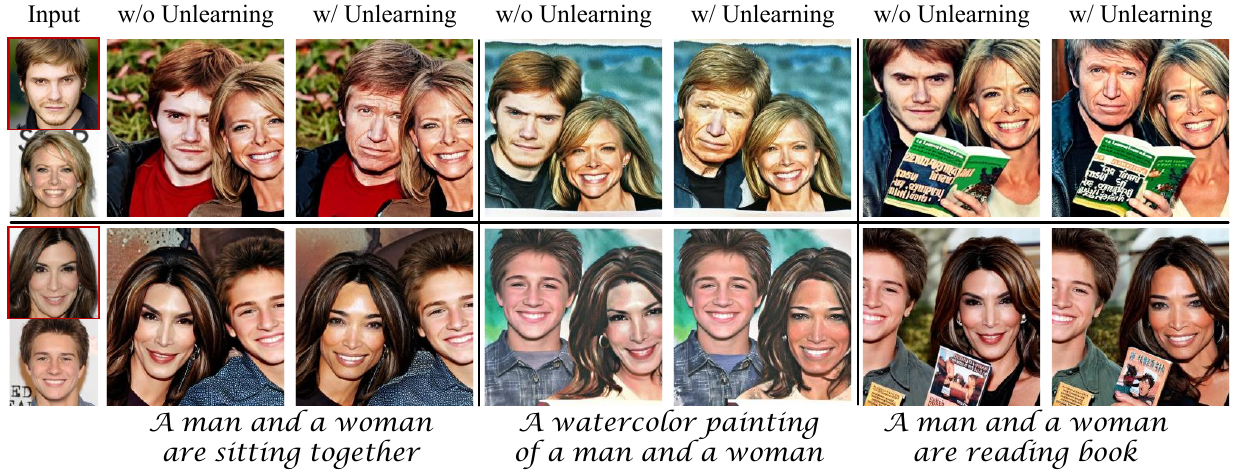} 
		\caption{
			Multi-subject image generation results of FastComposer~\cite{DBLP:journals/ijcv/XiaoYFDH25}.
			Our unlearned image encoder is adapted from the single-subject to multi-subject setting without any finetuning. For each example, the input image marked with a red box is the target identity to be unlearned, and the other one represents the identity to be retained.        
		}
		\label{fig:fast}
	\end{figure}
	
	\subsection{Generalization to New Generators}
	\begin{wraptable}{r}{0.6\textwidth}
		\vspace*{-3.5em}  
		\centering
		\setlength{\tabcolsep}{3pt}
		\caption{
			Quantitative evaluation of generalization ability when our unlearned image encoder is plugged into a new generator PhotoMaker~\cite{DBLP:conf/cvpr/LiCWQCS24} without any finetuning.}
		\label{tab:4}
		\begin{tabular}{l|ccc}
			\toprule
			Method & $\text{ID}_\text{target}^{g}$$(\downarrow)$ & $\text{ID}_\text{other}^{g}$$(\uparrow)$ & $\Delta$ID\\
			\midrule
			Baseline & 0.0872 & 0.1980 & 0.1108\\
			IREU  & 0.1365 & 0.8372 & 0.7007 \\
			\bottomrule
		\end{tabular}
		\vspace*{-2.5em}
	\end{wraptable}
	Since our method IREU only fine-tunes the image encoder while keeping the  diffusion model frozen, it can be plugged into new generators sharing a similar pipeline. This is a notable advantage compared with previous methods, e.g. BIA. 
	
	We first apply our unlearned image encoder to the new generator PhotoMaker~\cite{DBLP:conf/cvpr/LiCWQCS24} without any finetuning.  The quantitative results are reported in Table~\ref{tab:4}, where we can see our method IREU obtains a low ID similarity (0.1365 w.r.t. $\text{ID}_\text{target}^{g}$) for target identities to be unlearned and a high ID similarity (0.8372 w.r.t. $\text{ID}_\text{other}^{g}$) for other identities to be retained, reaching a good trade-off between unlearning and retaining (a high value of 0.7007 w.r.t. $\Delta$ID). Compared with the baseline which also obtains a low ID similarity (0.0872 w.r.t. $\text{ID}_\text{target}^{g}$) for target identities to be unlearned but a low ID similarity (0.1980 w.r.t. $\text{ID}_\text{other}^{g}$) for other identities to be retained, our method IREU shows a better generalization ability. These results indicate that our proposed identity-related feature perturbation enhances the robustness across different generators.
	We also show some qualitative results in Fig.~\ref{fig:photo}, where we can see the generated images for the target identity to be unlearned look rather different from the input image in terms of identity, while the generated images for the other identity to be retained well retain the identity compared to the input image.
	
	Furthermore, we apply our unlearned image encoder optimized based on the single-subject model of FastComposer~\cite{DBLP:journals/ijcv/XiaoYFDH25} to its multi-subject counterpart without any finetuning.
	As shown in Fig.~\ref{fig:fast}, our method is able to change the target identities to be unlearned and in the meantime retain the other identities.
	\begin{figure}[t]
		\centering
		\begin{subfigure}{0.45\linewidth}
			\centering
			\includegraphics[width=\linewidth]{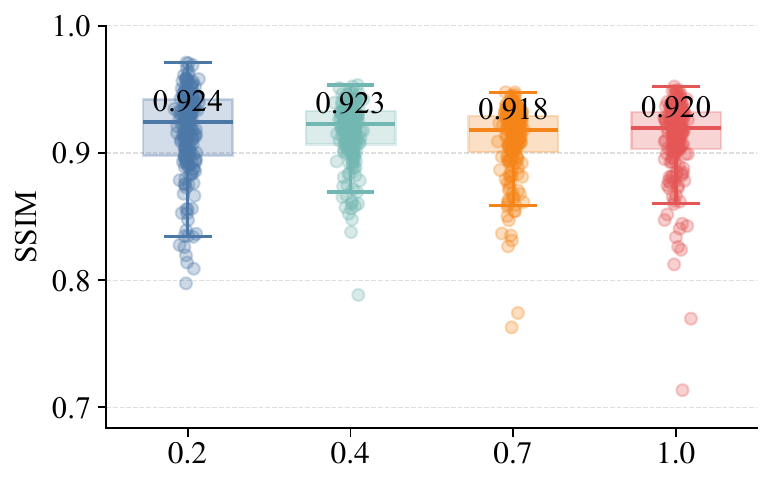}
			\caption{Boxplot of SSIM versus $k$.}
			\label{fig:sub_a}
		\end{subfigure}
		\hfill
		\begin{subfigure}{0.45\linewidth}    \includegraphics[width=\linewidth]{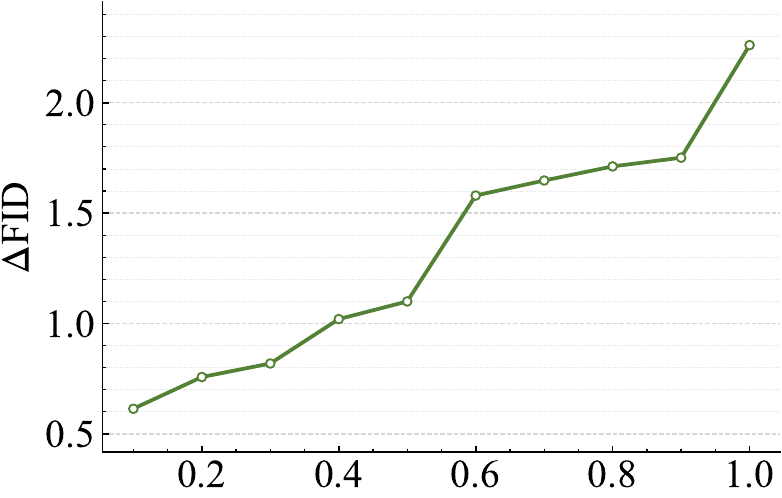}
			\caption{$\Delta\text{FID}$ values versus $k$.}
			\label{fig:sub_b}
		\end{subfigure}
		\hfill  
		\caption{
			Analysis on the ratio of identity-related features $k$.
			(a) The SSIM values show a larger variance (with more small values) as $k$ increases.
			(b) $\Delta\text{FID}$ consistently increases as $k$ grows.
			These results support our argument that restricting perturbations to identity-related elements preserves fidelity better.}
		\label{fig:parameter}
	\end{figure}
	
	\begin{figure}[t]
		\centering
		\includegraphics[width=0.9\columnwidth]{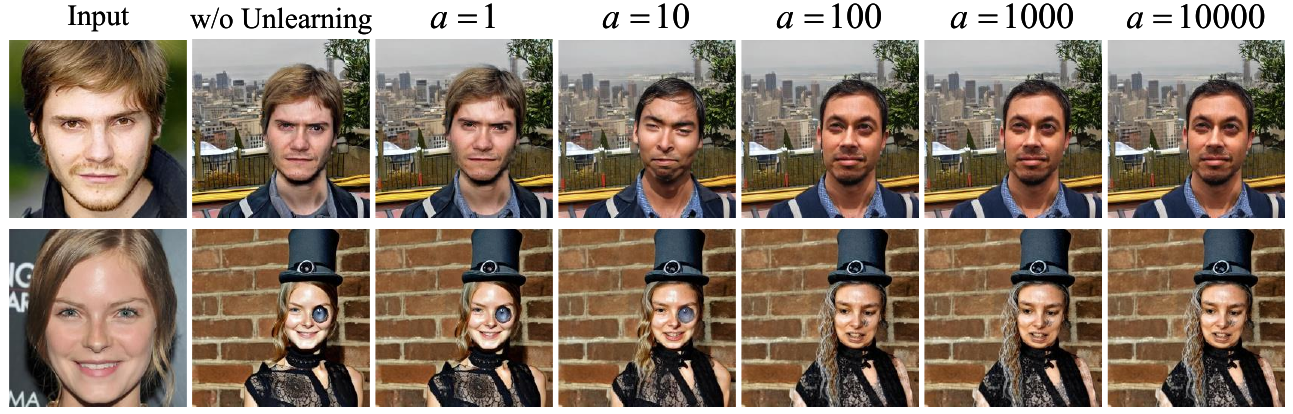} 
		\caption{
			Impact of the perturbation magnitude $\alpha$ in the identity-related transformation (Eq.~\ref{zu}).
			Increasing $\alpha$ consistently strengthens identity unlearning performance (lower ID similarity), but it saturates after $\alpha=100$.}
		\label{fig:alpha_ablation}
	\end{figure}
	\subsection{Ablation Study}
	\textbf{Impact of the ratio of identity-related features \(k\).}
	The ratio \(k\) in Eq.~\ref{m} determines the number of features that will be updated during unlearning for the target identities to be unlearned. Basically, the larger the value of \(k\), the more features will be updated, some of which may be identity-agnostic and are not expected to be updated. 
	As shown in Fig.~\ref{fig:parameter}, when \(k\) increases, more examples exhibit low SSIM values and the $\Delta\text{FID}$ value grows, both of which indicate that fidelity is decreasing. This phenomenon becomes more severe when \(k\) is larger than 0.4.
	Considering both unlearning performance and fidelity preservation, we choose $k=0.4$ in our experiments.
	
	\textbf{Impact of perturbation magnitude \(\alpha\).}
	The perturbation magnitude $\alpha$ in Eq.~\ref{zu} controls how much the identity-related features are perturbed given the mask.
	In Fig.~\ref{fig:alpha_ablation}, we analyze its effect on the generated outputs. 
	When $\alpha$ is small, the constructed virtual identity $\tilde{z}_{-}$ stays close to $z_{-}$, leading to insufficient unlearning.
	As $\alpha$ increases, the deviation from the input image grows and the identity similarity of the target identities to be unlearned quickly drops.
	Such a trend saturates after $\alpha = 100$, so we adopt $\alpha = 100$ in our experiments.

\begin{figure*}[t]
	\centering
	\begin{minipage}{0.48\textwidth}
		\centering
		\includegraphics[width=\linewidth]{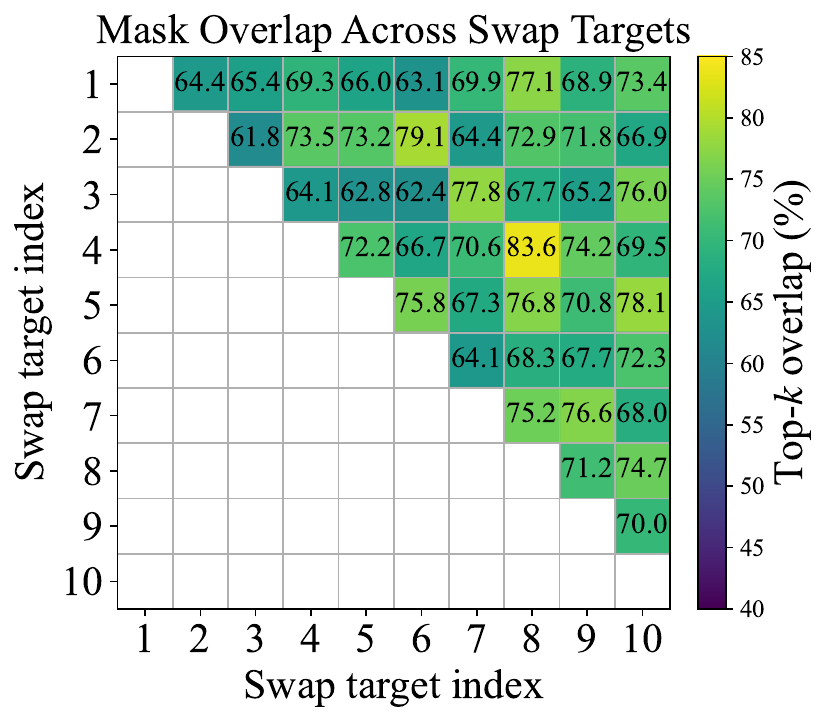}
		\caption{
			Impact of the randomly selected swap target \(x_a\).
			The mask overlap and final performance remain stable across different choices of \(x_a\).
		}
		\label{fig:swap}
	\end{minipage}
	\hfill
	\begin{minipage}{0.48\textwidth}
		\centering
		\includegraphics[width=\linewidth]{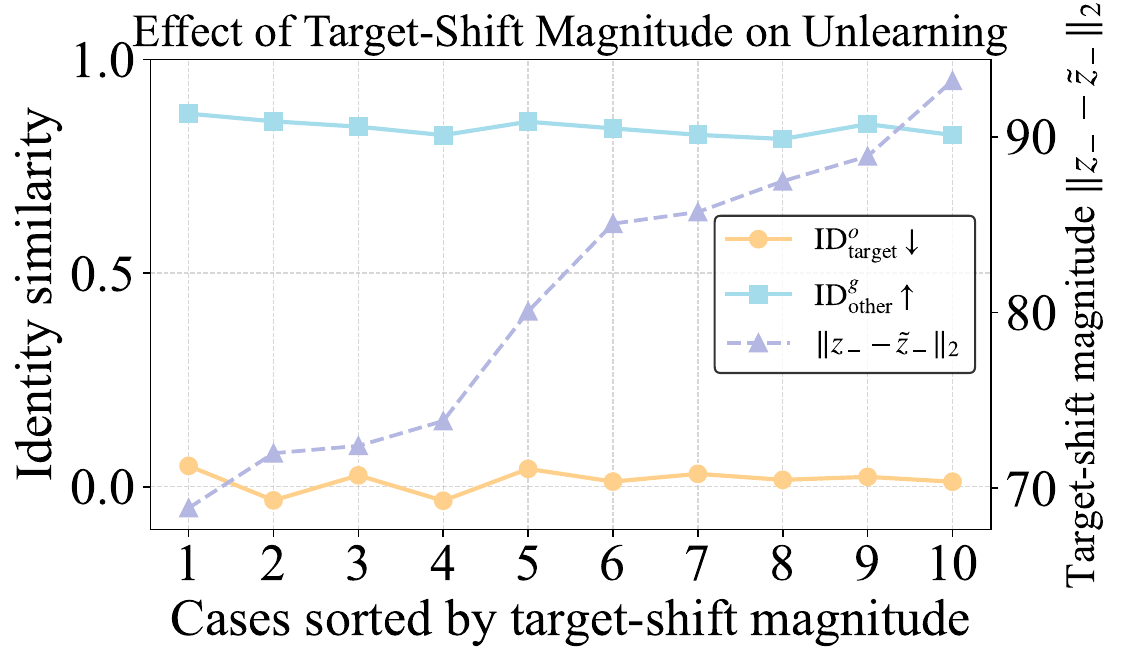}
		\caption{
			Analysis on the magnitude of masked identity-related differences.
			Cases are sorted by \(\|z_- - \tilde{z}_-\|_2\).
			The consistently low \(\text{ID}_\text{target}^{o}\) and high \(\text{ID}_\text{other}^{g}\) show that variations in masked differences do not harm the unlearning-retaining trade-off.
		}
		\label{fig:target_shift}
	\end{minipage}
\end{figure*}

	\textbf{Impact of the randomly selected swap target \(x_a\).}
	We analyze the influence of the swap target \(x_a\) used for constructing Face-Swap pairs. Specifically, for the same target identity \(x_-\), we sample 10 different \(x_a\) and evaluate the resulting mask \(\mathcal{M}\). As shown in Fig.~\ref{fig:swap}, the pairwise top-\(k\) mask overlap is consistently higher than the random expectation of \(40\%\) under \(k=0.4\), with a mean overlap of \(70.46\%\pm5.19\%\). This indicates that the selected identity-related dimensions are stable across different choices of \(x_a\).

	\textbf{Impact of masked identity-related difference magnitude.}
	We analyze how the target-shift magnitude affects unlearning stability during the unlearning process. Specifically, we sort different cases by \(\|z_- - \tilde{z}_-\|_2\), where \(\tilde{z}_-\) is constructed following Eq.~\ref{zu}. As shown in Fig.~\ref{fig:target_shift}, \(\text{ID}_\text{target}^{o}\) stays low and \(\text{ID}_\text{other}^{g}\) stays high across different target-shift magnitudes. These results demonstrate that mild variations in the identity-related shift do not degrade either identity unlearning or retaining performance.

	\section{Conclusion}
	Customized Portrait Generation (CPG) has become highly realistic and editable, while also posing significant privacy risks. Generative identity unlearning offers a crucial approach to address these privacy concerns, yet no specialized methods currently exist for identity unlearning in CPG.
	We present IREU, an Identity-Related Encoder-Only Unlearning framework for CPG.
	IREU achieves effective unlearning of target identities to be unlearned through its locating identity-related features and identity-related transformation components, while maintaining the fidelity of other identities to be retained.
	On CelebA-HQ, IREU achieves stronger identity suppression and better fidelity. The unlearned encoder can transfer to other CPG pipelines without retraining, demonstrating the effectiveness of encoder-only unlearning for CPG privacy protection.
	
	\section{Acknowledgements}
	This research is supported by the National Natural Science Foundation of China (Grant No. 62322602), and the Natural Science Foundation of Jiangsu Province, China (Grant No. BK20230033).
	\bibliographystyle{splncs04}
	\bibliography{main}
\end{document}